\documentclass[letterpaper, 10 pt, conference]{ieeeconf}  
\usepackage[utf8]{inputenc}

\IEEEoverridecommandlockouts                              

\usepackage{amsmath}
\usepackage{amssymb}
\usepackage{graphicx}
\usepackage{booktabs}
\usepackage{multirow}
\usepackage{url}
\usepackage{cite}
\usepackage{placeins}

\newcommand{\SEthree}{\mathrm{SE}(3)}

\newif\ifanonymoussubmission
\anonymoussubmissionfalse

\title{\LARGE \bf
WM-VS: Progress-Aligned World Models for Closed-Loop \\Visual Servoing
}

\ifanonymoussubmission
\author{}
\else
\author{
Guanzhong Sun$^{1}$,
Junyi Ma$^{2}$,
Yixuan Zhou$^{3}$,
Yuxuan Wu$^{2,4}$,
Yanzi Miao$^{1}$,
and Hesheng Wang$^{2,5}$%
\thanks{$^{1}$School of Information and Control Engineering, China University of Mining and Technology, Xuzhou, China.}%
\thanks{$^{2}$School of Automation and Intelligent Sensing, Shanghai Jiao Tong University, Shanghai, China.}%
\thanks{$^{3}$SJTU Global College, Shanghai Jiao Tong University, Shanghai, China.}%
\thanks{$^{4}$Shanghai Innovation Institute, Shanghai, China.}%
\thanks{$^{5}$Shanghai Key Laboratory of Navigation and Location Based Services, Shanghai Jiao Tong University, Shanghai, China.}%
\thanks{Corresponding author: Hesheng Wang.}%
}
\fi

\begin{document}

\maketitle
\thispagestyle{empty}
\pagestyle{empty}

\begin{abstract}
Closed-loop visual servoing requires predictions that indicate whether an action reduces task error, not only whether the action is plausible. We call this gap the \emph{prediction--control mismatch} and introduce WM-VS, a target-centric progress-aligned world-model framework for closed-loop visual servoing. Offline target-region DINOv2 correspondences define a signed four-dimensional servo coordinate for translation, scale, and in-plane rotation. Stage~1 aligns action-conditioned latent transitions with this coordinate; Stage~2 freezes the world model and trains a reactive joint-velocity policy with action imitation, consequence supervision, and short imagined rollouts that favor error contraction. Deployment is RGB-only and reactive, without online trajectory optimization. On a real 7-DoF eye-to-hand system, WM-VS reaches a corner RMSE no larger than 10\% of its initial value in 30/30 trials and retains this criterion at the final valid frame in 25/30 (83.33\%). Removing future-error alignment reduces retention to 26.67\%. The learned progress signal agrees with an external AprilTag corner error not used for training or control (mean Spearman $\rho=0.8778$). Without retraining, two unseen 3D targets achieve translation-error reductions of 86.48\% and 90.27\% and rotation-error reductions of 70.01\% and 65.70\%. These results link progress-aligned action consequences to repeated closed-loop correction and transfer.
Code and data will be released as open source.
\end{abstract}

\section{INTRODUCTION}

A learned visual servo controller must support repeated correction rather than only reproduce actions associated with individual observations. Each command changes the next observation, and the controller uses the updated feedback to compute the next command. Therefore, an action that appears plausible from the current observation may still produce drift if its consequence does not reduce the task-relevant visual error. The controller needs to evaluate not only what action is associated with an observation, but also whether the predicted consequence of that action moves the system toward the goal. We call the gap between action prediction and control-relevant progress the \emph{prediction--control mismatch}.

Classical VS resolves this issue explicitly. Image-based and position-based controllers define a task error and an interaction model that turns visual discrepancy into corrective motion \cite{hutchinson1996tutorial,chaumette2006visual,chaumette2007visual}. Direct and predictive variants broaden the usable measurements and horizons \cite{collewet2011photometric,allibert2010predictive}. Learning-based VS reduces dependence on hand-designed features by estimating pose, actions, or servoable representations from data \cite{bateux2018training,yu2019siamese,felton2021siamese3,felton2022autoencoder,felton2023dmlvs}.  However, many learned formulations primarily optimize state estimation or action prediction objectives, which indicate what action is associated with an observation but do not explicitly characterize how that action changes the task-relevant visual error under closed-loop execution.

This question becomes central with latent world models, which make action conditioned imagination a reusable abstraction for robot learning \cite{finn2017foresight,hirose2019deepvisualmpc,hafner2020dreamer,ma2026uni}. At the same time, self-supervised vision transformer (ViT) features provide dense spatial correspondences that transfer across appearance changes \cite{caron2021dino,oquab2024dinov2,amir2022deepvit}, and ViT-VS shows that such features can support explicit image-based servoing \cite{scherl2025vitvs}. Together, these developments motivate a different perspective: latent predictions should not only describe future states, but also organize candidate actions according to whether their consequences generate meaningful task progress in the closed loop.

WM-VS implements this idea by introducing a target-centric directional servo coordinate derived offline from target-region DINOv2 correspondences. The coordinate records signed image translation, relative scale, and in-plane rotation with respect to the desired visual goal, providing an explicit measure of object-relative progress. In Stage~1, the model learns action-conditioned latent transitions and aligns their predicted consequences with this directional servo coordinate. In Stage~2, the aligned world model is frozen and used to evaluate whether policy actions are expected to reduce target-centric visual error. Behavior cloning provides a data-supported action prior, while consequence supervision and short imagined rollouts encourage actions whose predicted outcomes contract the visual task error. The learned policy is then executed in closed loop and recomputes its command from current feedback at every control cycle.

This closed-loop perspective also changes evaluation. A controller may reach an accurate configuration while still producing actions that gradually increase error under continued feedback. We therefore distinguish \emph{reaching precision} from \emph{retaining precision}. Retention measures whether repeated feedback preserves the achieved accuracy and provides a direct behavioral evaluation of whether action consequences remain aligned with task progress.

\begin{figure*}[t]
\centering
\includegraphics[width=0.92\textwidth]{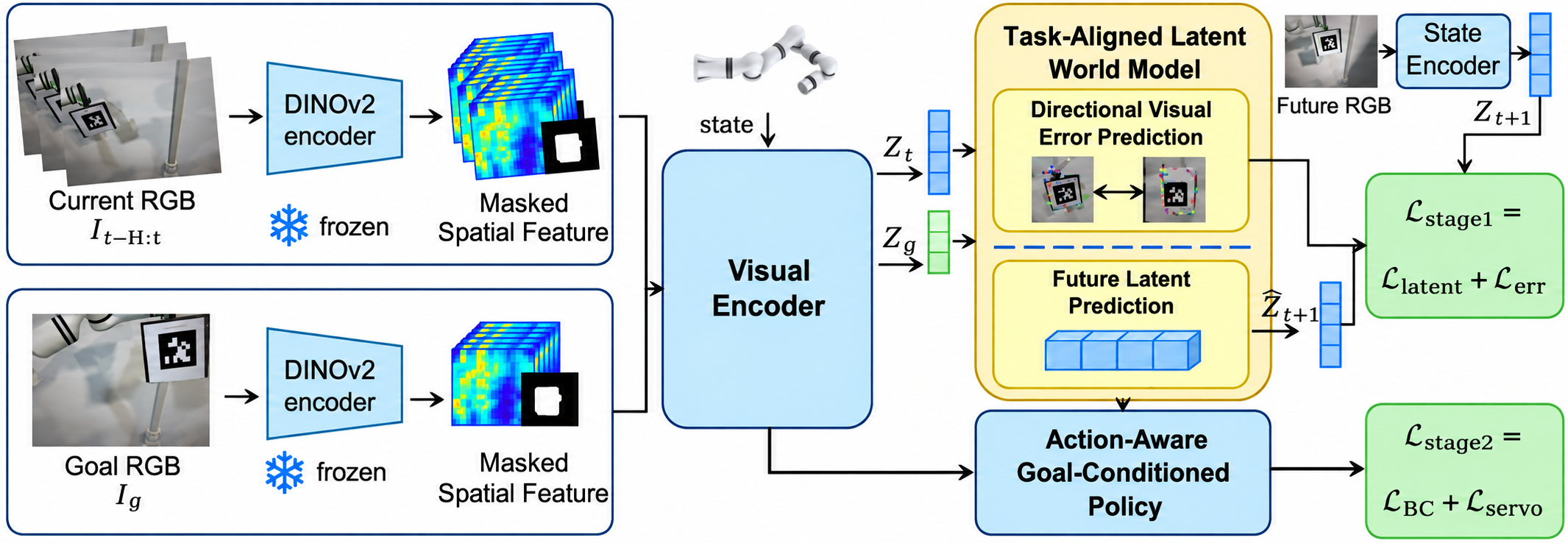}
\caption{\textbf{Overview of WM-VS.}
Frozen DINOv2 spatial features are focused on the task target and fused with robot history.
Stage~1 aligns action-conditioned latent transitions with a target-centric directional servo coordinate whose normalized reduction defines visual task progress.
Stage~2 freezes the aligned model and uses imagined action consequences to train a reactive goal-conditioned joint-velocity policy toward error contraction.}
\label{fig:overview}
\vspace{-0.5cm}
\end{figure*}

Our contributions are threefold.
\begin{itemize}
    \item We formulate the prediction--control mismatch in learned visual servoing and introduce a target-centric progress-aligned world model that evaluates predicted action consequences according to their effect on visual goal error.
    \item We use the frozen world model to train a reactive closed-loop policy that combines action imitation with consequence-aware supervision and short imagined rollouts, enabling goal-directed correction without online planning.
    \item We validate the representation to control link on a real 7-DoF system. WM-VS reaches the $10\%$ error region in 30/30 trials and retains it at the final valid frame in 25/30 trials. Future error alignment is directly tied to retention, the learned progress ordering agrees with an independent AprilTag signal, and zero shot tests on two new 3D targets show large TCP translation and rotation error reductions.
\end{itemize}

\section{RELATED WORK}

\subsection{Classical and Learning-Based Visual Servoing}

Classical VS maps visual error to corrective motion through interaction models \cite{hutchinson1996tutorial,chaumette2006visual,chaumette2007visual}. Variants use image moments \cite{chaumette2004moments}, direct photometric objectives \cite{collewet2008setfree,collewet2011photometric}, or predictive horizons \cite{lazar2009mpc,allibert2010predictive}. ViSP provides a generic software platform spanning a broad class of visual-servo control skills \cite{marchand2005visp}, while learning-based VS surveys cover neural-network, reinforcement-learning, and MPC-based formulations \cite{wu2022survey}. Learning-based VS regresses relative pose or actions \cite{bateux2018training,yu2019siamese,felton2021siamese3,saxena2017exploring,sampedro2018deeprl}, learns servoable representations \cite{lee2017learning,felton2022autoencoder,felton2023dmlvs}, or incorporates learned features and flow into geometric or predictive control \cite{puang2020kovis,adrian2022dfbvs,haugaard2022self,harish2020dfvs,katara2021deepmpcvs}. DeepMPCVS further couples learned optical-flow prediction with model-predictive control to optimize future motion \cite{katara2021deepmpcvs}. WM-VS instead learns explicit task-progress structure and requires neither an online interaction matrix nor deployment-time trajectory optimization.

\subsection{World Models for Control}

DINO/DINOv2 provide transferable dense descriptors for semantic correspondence \cite{caron2021dino,oquab2024dinov2,amir2022deepvit}. More broadly, Vision Transformers establish patch-token visual representations \cite{dosovitskiy2021vit}; Mask2Former and SAM provide general image-segmentation mechanisms for extracting object regions \cite{cheng2022mask2former,kirillov2023sam}, while Best-Buddies provides a robust criterion for template matching and correspondence selection \cite{dekel2015bestbuddies}. Together, these representation and matching tools support object-region extraction and correspondence, but do not themselves encode whether a candidate robot action produces progress toward the servo goal. ViT-VS uses DINOv2 correspondences online in an IBVS controller \cite{scherl2025vitvs}, whereas WM-VS uses them offline to supervise task progress. World-model methods exploit action-conditioned prediction through visual foresight \cite{finn2017foresight}, policy distillation \cite{hirose2019deepvisualmpc}, latent imagination \cite{hafner2020dreamer}, or predictive motion representations \cite{ma2026uni}. These works establish the value of prediction for control but do not explicitly organize transitions by visual-goal progress. WM-VS therefore uses its world model as a training-time consequence model rather than a deployment-time planner.

\section{METHOD}

\subsection{Problem Formulation and Target-Aware Representation}

We consider a fixed eye-to-hand camera observing a robot whose end-effector carries the task target.
At time $t$, the controller receives an RGB image $I_t$, joint state $\mathbf{q}_t$, recent joint and action history, and a desired RGB image $I_g$, and outputs a seven-joint velocity command $\dot{\mathbf{q}}_t$.
At every control cycle WM-VS refreshes the observation and robot state and recomputes the command, so it operates as a reactive closed-loop servo controller rather than replaying an open-loop action sequence.

Semantic object proposals from Mask2Former \cite{cheng2022mask2former} identify the task target in the initial and desired observations, and DINOv2 descriptors associate it across the two endpoints before causal tracking.
We use a frozen DINOv2 ViT-S/14 backbone with $448\times448$ input; normalized 384-D patch tokens from its final (12th) Transformer block form a $32\times32$ grid with patch size 14.
The implementation maps these descriptors to 32 visual channels and concatenates one target-mask channel, yielding a $32\times32\times33$ adapter input.
The adapter produces a 256-D visual embedding $\mathbf{v}_t$, with full-frame fallback for unreliable masks.
A causal temporal encoder fuses recent visual embeddings, joint states, previous actions, and control intervals into $\mathbf{z}_t$, while the desired image produces $\mathbf{z}_g$ through the same pathway.

The selected target regions are also used \emph{offline} to construct task-error supervision.
Within current and goal target masks, normalized DINOv2 patch descriptors are matched with a mutual-nearest-neighbor test related to best-buddy matching \cite{dekel2015bestbuddies}, low-similarity and ambiguous matches are rejected, and a random sample consensus (RANSAC) similarity transformation is estimated from the survivors.
No AprilTag measurement enters this offline supervision.

\subsection{Directional Servo Coordinate}

From robust target correspondences we define
\begin{equation}
\mathbf{e}_t =
[\Delta x_t,\Delta y_t,\Delta \log s_t,\Delta \theta_t]^\top ,
\label{eq:error}
\end{equation}
where $\Delta x$ and $\Delta y$ are current-minus-goal matched-centroid offsets in normalized image coordinates,
$\Delta \log s=\log(s_t/s_g)$,
and $\Delta \theta$ is the wrapped current-minus-goal in-plane rotation.

Because the four components have different natural scales, we compute a fixed component-wise scale vector $\boldsymbol{\sigma}$ on the training split and define
\begin{equation}
\tilde{\mathbf{e}}=\mathbf{e}\oslash\boldsymbol{\sigma},\qquad
d(\mathbf{e})=
\sqrt{\frac{1}{4}\|\tilde{\mathbf{e}}\|_2^2},
\label{eq:norm_error}
\end{equation}
where $\oslash$ denotes element-wise division, so that $d(\mathbf{e})$ is the root-mean-square magnitude of the normalized components.
We refer to $\mathbf{e}_t$ as a \emph{directional servo coordinate}.
The coordinate preserves translation, scale, and in-plane rotation differences that a scalar distance-to-goal would discard. It remains low-dimensional and is used only to organize task progress rather than as the controller state.
A decrease in $d(\mathbf{e})$ defines progress, while the policy itself continues to act on the richer spatial latent together with joint and action history.
Thus, the coordinate structures latent transitions without restricting the policy state to four dimensions.

\subsection{Progress-Aligned Latent Dynamics}

The purpose of the world model is not only to predict the next latent state, but also to make the consequence of an action measurable in the servo coordinate.
Given the current latent state and the executed robot action, the dynamics model predicts
\begin{equation}
\hat{\mathbf{z}}_{t+1}
=
f_{\theta}
(\mathbf{z}_t,\mathbf{q}_t,
\dot{\mathbf{q}}_t\Delta t,\Delta t),
\label{eq:dynamics}
\end{equation}
and a shared task-error head maps current and goal latents to the directional coordinate,
\begin{equation}
\hat{\mathbf{e}}_t
=
g_{\theta}(\mathbf{z}_t,\mathbf{z}_g),
\qquad
\hat{\mathbf{e}}_{t+1}
=
g_{\theta}(\hat{\mathbf{z}}_{t+1},\mathbf{z}_g).
\label{eq:error_head}
\end{equation}
The second relation is the key alignment constraint: the predicted next latent state must decode to the directional error induced by the action.
Consequently, latent transitions are organized not only by next-state prediction but also by whether their consequences represent progress, stagnation, or regression relative to the visual goal.
The Stage~1 objective is
\begin{equation}
\mathcal{L}_{\mathrm{WM}}
=
\mathcal{L}_{z}
+
\lambda_e\mathcal{L}_{e}
(\hat{\mathbf{e}}_t,\mathbf{e}_t)
+
\lambda_f\mathcal{L}_{e}
(\hat{\mathbf{e}}_{t+1},\mathbf{e}_{t+1}),
\label{eq:stage1}
\end{equation}
where $\mathcal{L}_{z}$ combines smooth-$L_1$ next-latent regression and cosine consistency, and $\mathcal{L}_e$ is smooth-$L_1$ regression in the fixed normalized coordinates of (\ref{eq:norm_error}).
We use $\lambda_e=1$ and $\lambda_f=0.5$.
This objective turns predictive latent dynamics into a servo-centric world model: the latent retains appearance and robot-history information, while action-conditioned transitions are explicitly ordered by task progress. The model is therefore optimized for control-relevant consequence structure rather than image synthesis.

\subsection{Action-Consequence-Aware Policy Learning}

After Stage~1, the aligned world model is frozen; we denote its fixed parameters by $\bar{\theta}$ and reuse it as a differentiable model of action consequences.
A goal-conditioned Transformer receives the current latent, goal latent, their latent difference, and recent joint history, and predicts
\begin{equation}
\dot{\mathbf{q}}_t^{\pi}
=
\pi_{\psi}
(\mathbf{z}_t,\mathbf{z}_g,\mathbf{q}_{t-h:t}).
\label{eq:policy}
\end{equation}

\begin{figure}[t]
\centering
\includegraphics[width=\linewidth]{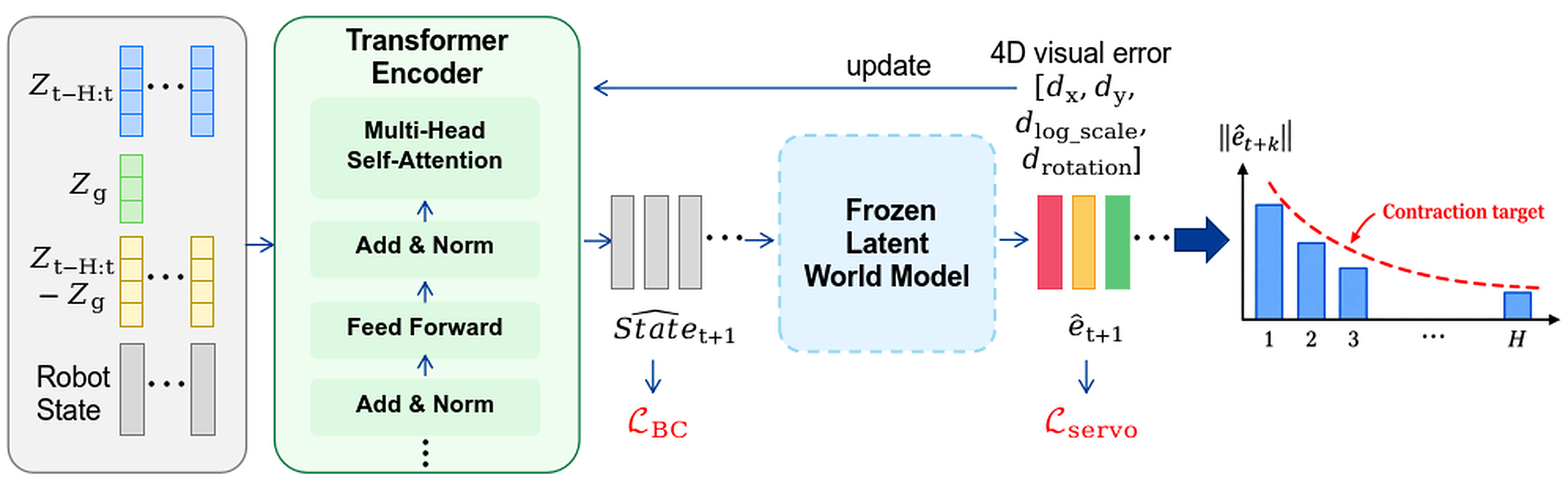}
\caption{\textbf{Directional servo coordinate for action-consequence learning.}
The goal-conditioned policy predicts a joint action from the current and goal latents together with robot history.
The aligned predictive representation maps the imagined consequence to the four-dimensional error coordinate.
Action imitation constrains the policy near the data distribution, while the servo terms regularize local consequences and encourage short-horizon error contraction.}
\label{fig:policy_learning}
\vspace{-0.5cm}
\end{figure}

Fig.~\ref{fig:policy_learning} shows how the directional servo coordinate links a policy action to its predicted visual consequence.

Let $\dot{\mathbf{q}}_t^{D}$ denote the logged joint-velocity action and
$\mathbf{e}_{t+1}^{D}$ the error observed after its execution.
The imitation term is
\begin{equation}
\mathcal{L}_{\mathrm{BC}}
=
\ell_a(\dot{\mathbf{q}}_t^{\pi},\dot{\mathbf{q}}_t^{D}).
\label{eq:bc}
\end{equation}

The policy action is propagated through the frozen world model and locally aligned with the demonstrated consequence,
\begin{equation}
\begin{aligned}
\hat{\mathbf{z}}^\pi_{t+1}
&=
f_{\bar{\theta}}\!\left(
\mathbf{z}_t,\mathbf{q}_t,
\dot{\mathbf{q}}^\pi_t\Delta t_t,\Delta t_t
\right),\\
\hat{\mathbf{e}}^\pi_{t+1}
&=
g_{\bar{\theta}}\!\left(
\hat{\mathbf{z}}^\pi_{t+1},\mathbf{z}_g
\right),\\
\mathcal{L}_{\mathrm{next}}
&=
\operatorname{MaskedMean}\!\Bigl[\\[-0.35em]
&\quad
\operatorname{SmoothL1}\!\left(
\hat{\mathbf{e}}^\pi_{t+1}\oslash\boldsymbol{\sigma},
\mathbf{e}^{D}_{t+1}\oslash\boldsymbol{\sigma}
\right);
\,m_t m_{t+1}
\Bigr].
\end{aligned}
\label{eq:next}
\end{equation}
where $m_t\in\{0,1\}$ denotes the validity of the directional-error target at step $t$, so $m_t m_{t+1}$ retains only transitions with valid supervision at both steps.

Because $\dot{\mathbf{q}}^\pi_t$ need not equal the demonstrated action $\dot{\mathbf{q}}^D_t$, $\mathbf{e}^{D}_{t+1}$ is not an exact on-policy outcome target.
 Under the behavior-cloning constraint, $\mathcal{L}_{\mathrm{next}}$ instead provides local consequence alignment around demonstrated actions; policy-specific error reduction is encouraged by $\mathcal{L}_{\mathrm{contract}}$ through imagined rollouts.

The aligned world model also allows the policy's own actions to be evaluated. Because the predicted next latent states carry the servo coordinate, short policy-induced rollouts can be regularized according to whether they reduce task error.
Starting from $\mathbf{z}_t$, the policy and frozen world model are recursively applied for a short horizon $K$, and for the imagined policy-induced task errors
$\hat{\mathbf{e}}_{t+k}^{\pi}$ we penalize
\begin{equation}
\mathcal{L}_{\mathrm{contract}}
=
\frac{1}{K}
\sum_{k=1}^{K}
\left[
d(\hat{\mathbf{e}}_{t+k}^{\pi})
-
\left(
\rho^k d(\mathbf{e}_t)+\epsilon
\right)
\right]_+ ,
\label{eq:contraction}
\end{equation}
where $[x]_+=\max(0,x)$.
This term needs no observed next-state target for the policy's imagined action; it directly asks the policy-induced rollout to reduce task error.
A small near-goal action penalty $\mathcal{L}_{\mathrm{near}}$ suppresses unnecessary motion close to the target, and the complete policy objective is
\begin{equation}
\mathcal{L}_{\pi}
=
\mathcal{L}_{\mathrm{BC}}
+
\lambda_n\mathcal{L}_{\mathrm{next}}
+
\lambda_c\mathcal{L}_{\mathrm{contract}}
+
\lambda_0\mathcal{L}_{\mathrm{near}}.
\label{eq:stage2}
\end{equation}
We use $\lambda_n=1$, $\lambda_c=1$, and $\lambda_0=0.05$.
The three terms are complementary: $\mathcal{L}_{\mathrm{BC}}$ anchors demonstrated actions, $\mathcal{L}_{\mathrm{next}}$ supervises local consequences, and $\mathcal{L}_{\mathrm{contract}}$ rewards predicted error reduction. Stage~2 thus trains for error-contracting closed-loop behavior rather than one-step action accuracy alone.

\subsection{Training and Deployment}
\label{sec:impl}

The visual adapter produces 256-D embeddings by summing a fixed anchor branch ($8\times8$ adaptive pooling and fixed random projection) and a trainable residual branch (depthwise/pointwise convolutions with GroupNorm/GELU, reducing $32\times32$ to $8\times8$, followed by a linear projection), then applying LayerNorm. The temporal encoder and Stage~2 policy each use 8 Transformer layers, 8 heads, and a 512-D feed-forward width; the latent dimension is 256 and $h=10$ previous macro-keyframes.

We use $K=3$, $\rho=0.9$, and $\epsilon=0.02$. All normalization statistics in (\ref{eq:norm_error}) are computed from the training split and then fixed. Both stages are trained for at most 5000 epochs with early-stopping patience 200.

At deployment, the learned policy is updated at approximately $5\,\mathrm{Hz}$ from the latest RGB observation and robot feedback. Each policy output is interpolated by the low-level interface into smooth joint commands sent at up to $30\,\mathrm{Hz}$. A bounded gain derived from predicted error attenuates command magnitude near the goal. Perception uses causal target tracking with fallback handling. The controller uses no depth, AprilTag measurement, future observation, online explicit feature geometry, latent-rollout optimization, or trajectory planning.

\section{EXPERIMENTS}

\subsection{Platform, Data, and Evaluation Protocol}

We use the suffix E2H to denote the fixed-camera eye-to-hand adaptations of the corresponding methods. Experiments use a fixed eye-to-hand Gemini~2 RGB-D camera and a seven-axis RealMan RMC-DA robot. The RGB stream is $640\times480$ (4:3), with horizontal, vertical, and diagonal fields of view of $63.55^\circ$, $49.82^\circ$, and $75.49^\circ$, respectively. A custom-designed, 3D-printed end-effector is rigidly bolted to the robot flange and carries the task target. Although the sensor provides RGB-D, WM-VS consumes RGB only; registered depth is used only by the ViT-VS-E2H baseline.

For each data-collection trajectory, two joint configurations are randomly sampled within a predefined safe joint-space region and retained only when all AprilTag corners are detectable at both endpoints. One endpoint is randomly selected as the goal; after a $3$-s settling period, the desired image $I_g$ is captured. The robot then moves to the other endpoint, settles for another $3$~s, and captures the initial image. During the subsequent trajectory, RGB images, joint positions, and joint velocities are logged at $5\,\mathrm{Hz}$.

The resulting dataset contains 1000 trajectories and is split \emph{by trajectory} into training, validation, and test partitions in a $70{:}15{:}15$ ratio (700/150/150 trajectories). This prevents temporally adjacent frames from the same trajectory from crossing partitions, and all normalization statistics used by the directional servo coordinate are estimated from the training partition. Real-robot evaluation uses 30 independently executed trials per method or ablation. Test trials start with large displacements relative to the desired pose: the initial corner RMSE is approximately 400--500~px, translation discrepancy is 300--400~mm, and rotation discrepancy spans $19^\circ$--$61^\circ$.

An OpenCV DICT\_APRILTAG\_36h11 marker \cite{olson2011apriltag}
(ID~0, side length $0.066125$~m), printed on A4 paper, provides the common evaluation signal.
For WM-VS, tag detection is used only to screen data-collection endpoints for target visibility and to provide the external evaluator and stopping supervisor; no tag-derived quantity enters the visual encoder, latent dynamics, policy, training losses, or online gain.

Moment-VS is the sole controller that uses the detected tag geometry online, as detailed in~\ref{sec:baselines}.

Let the four detected corners in the current image be
$\mathbf{c}_i=(u_i,v_i)$ and the corresponding corners in the desired image be
$\mathbf{c}_i^{*}=(u_i^{*},v_i^{*})$, with consistent corner ordering and $i=1,\ldots,4$.
The image-space evaluation metric is the corner RMSE
\begin{equation}
E_{\mathrm{corner}}(t)
=
\sqrt{
\frac{1}{4}
\sum_{i=1}^{4}
\|\mathbf{c}_i-\mathbf{c}_i^{*}\|_2^2
}
\label{eq:corner}
\end{equation}
in pixels, which quantifies image-space deviation between the current and desired tag quadrilaterals.

The stopping supervisor and external progress-consistency analysis use a separate, dimensionless normalized corner error.
With camera intrinsics $(f_x,f_y,c_x,c_y)$, each detected pixel corner is normalized as
\begin{equation}
\bar{\mathbf{c}}_i =
\begin{bmatrix}
(u_i-c_x)/f_x\\
(v_i-c_y)/f_y
\end{bmatrix},
\qquad
\bar{\mathbf{c}}_i^{*} =
\begin{bmatrix}
(u_i^{*}-c_x)/f_x\\
(v_i^{*}-c_y)/f_y
\end{bmatrix},
\end{equation}
and we define
\begin{equation}
E_{\mathrm{tag}}(t)
=
\sqrt{
\sum_{i=1}^{4}
\left\|
\bar{\mathbf{c}}_i-\bar{\mathbf{c}}_i^{*}
\right\|_2^2
}.
\label{eq:tag_norm}
\end{equation}
We denote this normalized image-space corner error by $E_{\mathrm{tag}}$; it is neither a TCP pose error nor a metric in millimeters or degrees.

For additional reporting, the known tag size and camera intrinsics are used with perspective-n-point (PnP) to recover current and desired AprilTag poses. Translation error is the Euclidean distance between the two tag translations, reported in millimeters, and rotation error is the geodesic angle of their relative rotation, reported in degrees. These quantities describe AprilTag pose in the camera frame and are kept separate from the normalized image-space metric $E_{\mathrm{tag}}$.

Each method and ablation variant is evaluated in 30 real-robot trials.
We report whether a run ever reaches
\begin{equation}
\min_t E_{\mathrm{corner}}(t)
\le
0.1\,E_{\mathrm{corner}}(0)
\label{eq:convergence_region}
\end{equation}
and whether the final valid frame satisfies the same $10\%$ error criterion. This criterion is defined relative to each trial's initial error and is used only for evaluation.
Thus trajectories are not truncated at first entry and can expose whether subsequent closed-loop feedback preserves or loses the reached precision.
All methods share the same external stopping supervisor. A trial terminates when $E_{\mathrm{tag}}$ does not decrease for 50 consecutive closed-loop updates or when tag corners are unavailable for 30 consecutive frames, in which case the last valid frame is used.

For trials whose final valid frame remains inside the reporting region, we report the relative post-minimum rebound
$R_{\mathrm{rebound}}=[E_{\mathrm{corner}}(T)-E_{\mathrm{best}}]/E_{\mathrm{best}}$,
where $E_{\mathrm{best}}=\min_t E_{\mathrm{corner}}(t)$.
Let $t_\star$ denote the best-observed step. Over the first 50 controller updates after $t_\star$, we additionally report the fractions satisfying
$E_{\mathrm{corner}}(t)\leq E_{\mathrm{corner}}(t-1)$
and
$E_{\mathrm{corner}}(t)<E_{\mathrm{corner}}(t-1)$,
denoted \emph{non-increase} and \emph{strict decrease}.
A step denotes one closed-loop update of the corresponding controller; these update-normalized statistics are scale-free and complement final retention and rebound.

\begin{table*}[t]
\vspace{-0.5cm}
\caption{Compared methods and online information used by the controller.}
\label{tab:methods}
\centering
\small
\setlength{\tabcolsep}{6pt}
\begin{tabular}{lcccc}
\toprule
Method & Task-specific training & RGB & Depth & Online explicit features\\
\midrule
WM-VS (Ours) & \checkmark & \checkmark & -- & --\\
WM-VS w/o Future-Error Alignment & \checkmark & \checkmark & -- & --\\
WM-VS w/o Error Guidance & \checkmark & \checkmark & -- & --\\
WM-VS (BC-only) & \checkmark & \checkmark & -- & --\\
VSNet-E2H & \checkmark & \checkmark & -- & --\\
Moment-VS & -- & \checkmark & -- & $\checkmark^{\dagger}$\\
ViT-VS-E2H & -- & \checkmark & $\checkmark^{\ddagger}$ & \checkmark\\
\bottomrule
\addlinespace[2pt]
\multicolumn{5}{l}{\footnotesize $^{\dagger}$Plane parameters $(A,B,C)$ of the model $1/Z=Ax+By+C$ obtained online from the marker pose.}\\
\multicolumn{5}{l}{\footnotesize $^{\ddagger}$Registered depth from the RGB-D sensor.}\\
\end{tabular}
\vspace{-0.5cm}
\end{table*}

\subsection{Baselines and Eye-to-Hand Adaptation}
\label{sec:baselines}

Table~\ref{tab:methods} summarizes the online information available to each controller. WM-VS uses RGB observations and robot proprioception without depth, target-pose estimation, an online interaction matrix, or trajectory optimization. The geometric baselines retain the sensing required by their original control formulations.

ViT-VS-E2H.
We reproduce the public ViT-VS implementation \cite{scherl2025vitvs}, preserving DINOv2 feature extraction, local correspondence matching, interaction-matrix construction, and visual-servo gain computation.
The original method is eye-in-hand, so we adapt only the sensing and coordinate interfaces: fixed-camera RGB and registered depth are provided to the original visual-servo computation, the resulting camera-frame velocity is transformed to the robot base using the eye-to-hand calibration, and the robot spatial Jacobian pseudoinverse maps end-effector velocity to joint velocity.

Moment-VS.
We implement image-moment visual servoing with ViSP \cite{chaumette2004moments,marchand2005visp}. At each trial, desired moment features are initialized from the AprilTag corners and PnP pose at the goal configuration. During servoing, detected corners define the normalized tag polygon and the moment features $(x_g,y_g,a_n,s_x,s_y,\alpha)$. AprilTag PnP provides the online plane model $1/Z=Ax+By+C$, whose parameters are recomputed every frame. ViSP uses gain $\lambda=0.4$; the resulting camera velocity is transformed to the robot base and mapped to joint velocity through the robot Jacobian.

VSNet-E2H.
We reproduce the VSNet design \cite{yu2019siamese}, retaining the Siamese AlexNet architecture, paired current and desired RGB input, and joint translation and quaternion regression, with training labels from RealMan forward kinematics rather than AprilTag PnP.
At runtime, the predicted relative TCP pose becomes a bounded $\SEthree$ increment and a proportional end-effector command that is mapped to joint motion through the Jacobian, with all predicted quantities and per-step joint increments bounded before execution.

\subsection{Reaching the Goal Neighborhood}

\begin{figure}[t]
\centering
\setlength{\tabcolsep}{2pt}
\begin{tabular}{cc}
\includegraphics[width=0.48\linewidth]{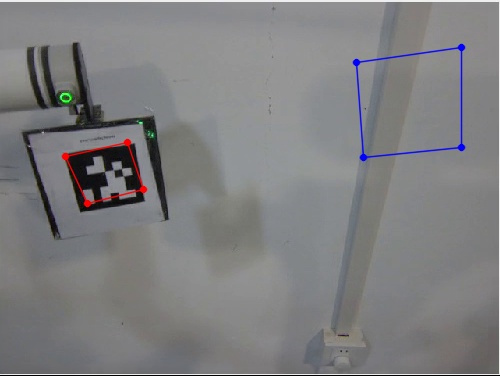} &
\includegraphics[width=0.48\linewidth]{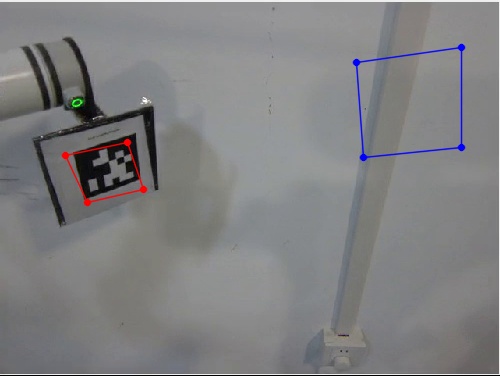} \\
{\small (a) WM-VS (Ours)} &
{\small (b) VSNet-E2H} \\[2pt]
\includegraphics[width=0.48\linewidth]{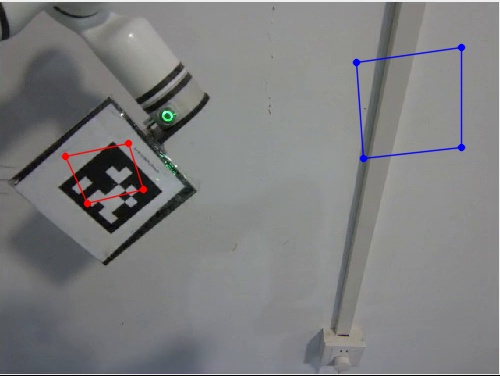} &
\includegraphics[width=0.48\linewidth]{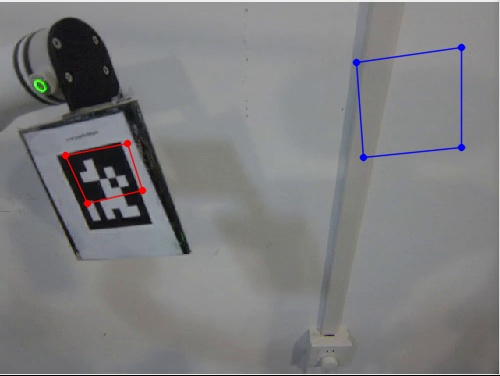} \\
{\small (c) Moment-VS} &
{\small (d) ViT-VS-E2H}
\end{tabular}
\caption{\textbf{Best-observed real-robot precision.}
(a) WM-VS, (b) VSNet-E2H, (c) Moment-VS, and (d) ViT-VS-E2H under identical initial/desired configurations. Blue/red quadrilaterals mark initial/desired AprilTag corners.}
\label{fig:precision_examples}
\vspace{-0.5cm}
\end{figure}

We first ask whether each controller can reach a precise configuration at least once. Fig.~\ref{fig:precision_examples} shows best-observed frames under identical initial and desired joint configurations, and Table~\ref{tab:results} summarizes 30-trial convergence and best-observed errors.

\begin{table*}[t]
\caption{Transient accuracy, final retention, and post-minimum local behavior over 30 real-robot trials.}
\vspace{-0.25cm}
\label{tab:results}
\centering
\scriptsize
\setlength{\tabcolsep}{1.5pt}
\renewcommand{\arraystretch}{0.95}
\begin{tabular*}{\textwidth}{@{\extracolsep{\fill}}lcccccc@{}}
\toprule
Method
& Conv.$^{\dagger}$
& Best RMSE (px)$^{\S}$
& Best trans. (mm)$^{\S}$
& Best rot. ($^\circ$)$^{\S}$
& Final ret.$^{\ddagger}$
& Post-min. 50-step non-inc./dec. (\%)\\
\midrule
WM-VS (Ours)
& \textbf{100.00\%}
& {\boldmath$10.20\pm2.30$}
& $13.21\pm2.32$
& {\boldmath$5.78\pm2.55$}
& \textbf{83.33\%}
& $63.3/34.0$\\

WM-VS w/o Future-Error Alignment
& 86.67\%
& $12.16\pm6.51$
& $20.78\pm9.51$
& $6.49\pm3.82$
& 26.67\%
& $65.6/26.3$\\

WM-VS w/o Error Guidance
& 36.67\%
& $13.51\pm5.16$
& $23.06\pm14.05$
& $9.09\pm4.53$
& 23.33\%
& $40.3/16.7$\\

WM-VS (BC-only)
& 90.00\%
& $11.44\pm8.42$
& {\boldmath$12.86\pm3.55$}
& $6.61\pm3.25$
& 50.00\%
& $68.8/20.8$\\
\midrule

VSNet-E2H
& \textbf{100.00\%}
& $13.18\pm2.17$
& $14.55\pm4.12$
& $27.11\pm4.31$
& 0.00\%
& $2.5/2.0$\\

Moment-VS
& 66.67\%
& $20.47\pm11.55$
& $25.52\pm15.93$
& $18.11\pm7.29$
& 56.67\%
& $49.3/39.8$\\

ViT-VS-E2H
& 60.00\%
& $29.64\pm3.12$
& $79.24\pm65.40$
& $48.10\pm7.96$
& 60.00\%
& $62.4/33.2$\\
\bottomrule
\addlinespace[2pt]
\multicolumn{7}{p{0.985\textwidth}}{\scriptsize
$^{\dagger}$Best corner RMSE $\leq 0.1E_{\mathrm{corner}}(0)$.
$^{\ddagger}$Final valid frame satisfies the same criterion.
$^{\S}$Best-observed errors are computed over converged trials.
Post-min. reports, over the first 50 closed-loop updates after each trajectory minimum, the percentages with
$E_t\leq E_{t-1}$ / $E_t<E_{t-1}$.
}\\
\end{tabular*}
\vspace{-0.5cm}
\end{table*}

WM-VS reaches the $10\%$ region in all 30 trials, with best-observed errors of $10.20\pm2.30$~px, $13.21\pm2.32$~mm, and $5.78\pm2.55^\circ$ in corner RMSE, translation, and rotation, respectively. VSNet-E2H also reaches the region in every trial but with markedly larger rotation error; Moment-VS and ViT-VS-E2H converge in $66.67\%$ and $60.00\%$. Thus, progress-aligned training preserves strong initial convergence. The remaining question is whether this precision is maintained after the controller first reaches the goal region.

\subsection{Retaining Precision Under Continued Feedback}

\begin{figure}[t]
\centering
\includegraphics[width=\linewidth]{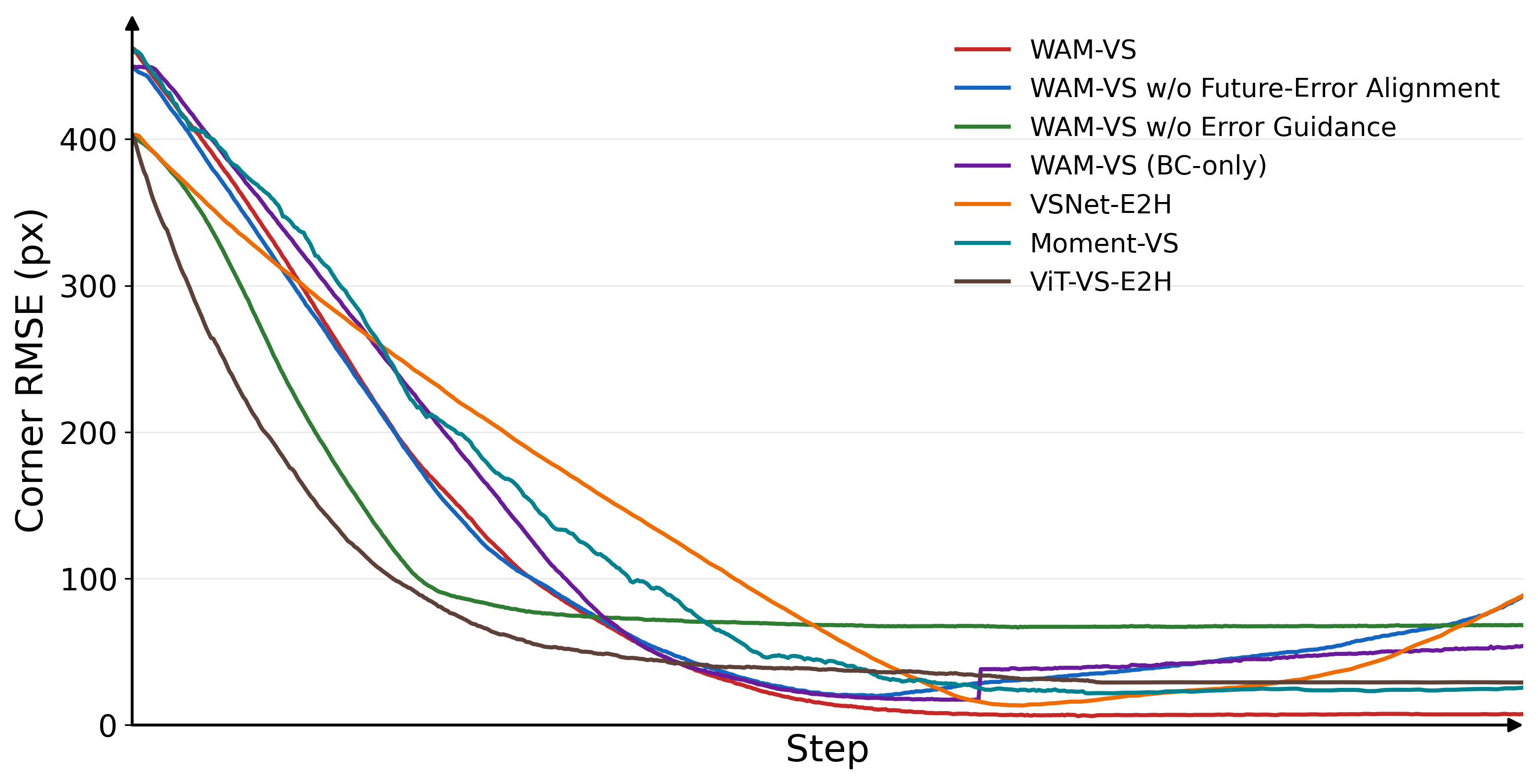}
\caption{\textbf{Representative closed-loop corner-RMSE trajectories.}
Trajectories continue beyond their minima to the final valid step, exposing post-minimum rebound that best error alone does not capture.}
\label{fig:error_curves}
\vspace{-0.5cm}
\end{figure}

Best-observed accuracy can hide a controller that reaches the goal neighborhood and then drifts away. We therefore plot trajectories through the final valid step in Fig.~\ref{fig:error_curves} and evaluate both final retention and scale-free local behavior after the trajectory minimum.

WM-VS satisfies the $10\%$ error criterion at the final valid frame in
25/30 (83.33\%) trials.
Across these 25 trials, the final corner RMSE is
$10.36\pm2.38$~px, and the mean relative rebound from the trajectory
minimum is only $0.0870\pm0.0936$.
The comparison with VSNet-E2H shows why final retention is a closed-loop measure rather than a best-frame measure. Both methods reach the $10\%$ criterion in all 30 trials, but VSNet-E2H satisfies it at the final valid frame in 0/30 trials, with every trajectory ending at least four times above its own best corner RMSE.
Moment-VS and ViT-VS-E2H satisfy the final criterion in 56.67\% and
60.00\% of trials, respectively.
In the evaluated RGB-only eye-to-hand setting, WM-VS therefore provides
the strongest combination of reaching performance and near-goal retention.

\subsection{What Creates the Retention Advantage?}

Three ablations isolate the progress-alignment pathway. w/o Future-Error Alignment sets $\lambda_f=0$ while retaining current-error supervision and all Stage~2 objectives. BC-only retains the full Stage~1 model but removes $\mathcal{L}_{\mathrm{next}}$, $\mathcal{L}_{\mathrm{contract}}$, and $\mathcal{L}_{\mathrm{near}}$. w/o Error Guidance removes current/future error supervision, all servo-aware Stage~2 losses, and predicted-error gain scaling while retaining latent prediction and behavior cloning.

Removing only future-error alignment preserves $86.67\%$ convergence but drops final retention from $83.33\%$ to $26.67\%$. BC-only reaches the region in $90.00\%$ of trials and retains it in $50.00\%$. The post-minimum sign statistics provide complementary evidence: the full model has the highest strict-decrease rate among the WM-VS variants ($34.00\%$), while BC-only has a higher non-increase rate ($68.80\%$) but substantially fewer strictly decreasing steps ($20.80\%$). We therefore treat these sign statistics as a local feedback diagnostic rather than a standalone stability score. Because w/o Future-Error Alignment and BC-only retain the same online gain rule as the full model, the retention gap cannot be attributed to gain scaling alone. Removing the entire error-guidance pathway reduces convergence/retention to $36.67\%/23.33\%$. These ablations support distinct roles for the two forms of guidance: current-error guidance mainly supports reaching, whereas future-error alignment and action-consequence learning improve behavior after entering the goal neighborhood.

\subsection{External Progress Consistency}

We next test whether the learned progress ordering agrees with the external normalized AprilTag corner error $E_{\mathrm{tag}}$ in (\ref{eq:tag_norm}), which is never used as a WM-VS training target or controller input. For this analysis only, both sequences are smoothed with the same five-frame moving average.

Across 30 trajectories, the learned signal achieves mean Pearson $r=0.9911$ and mean Spearman $\rho=0.8778$ with $E_{\mathrm{tag}}$. Spearman correlation is more relevant here because it measures progress ordering. The result indicates that the learned servo coordinate preserves the monotonic structure of the external image-space error without receiving that signal during training or control.

\begin{figure}[t]
\centering
\includegraphics[width=0.75\linewidth]{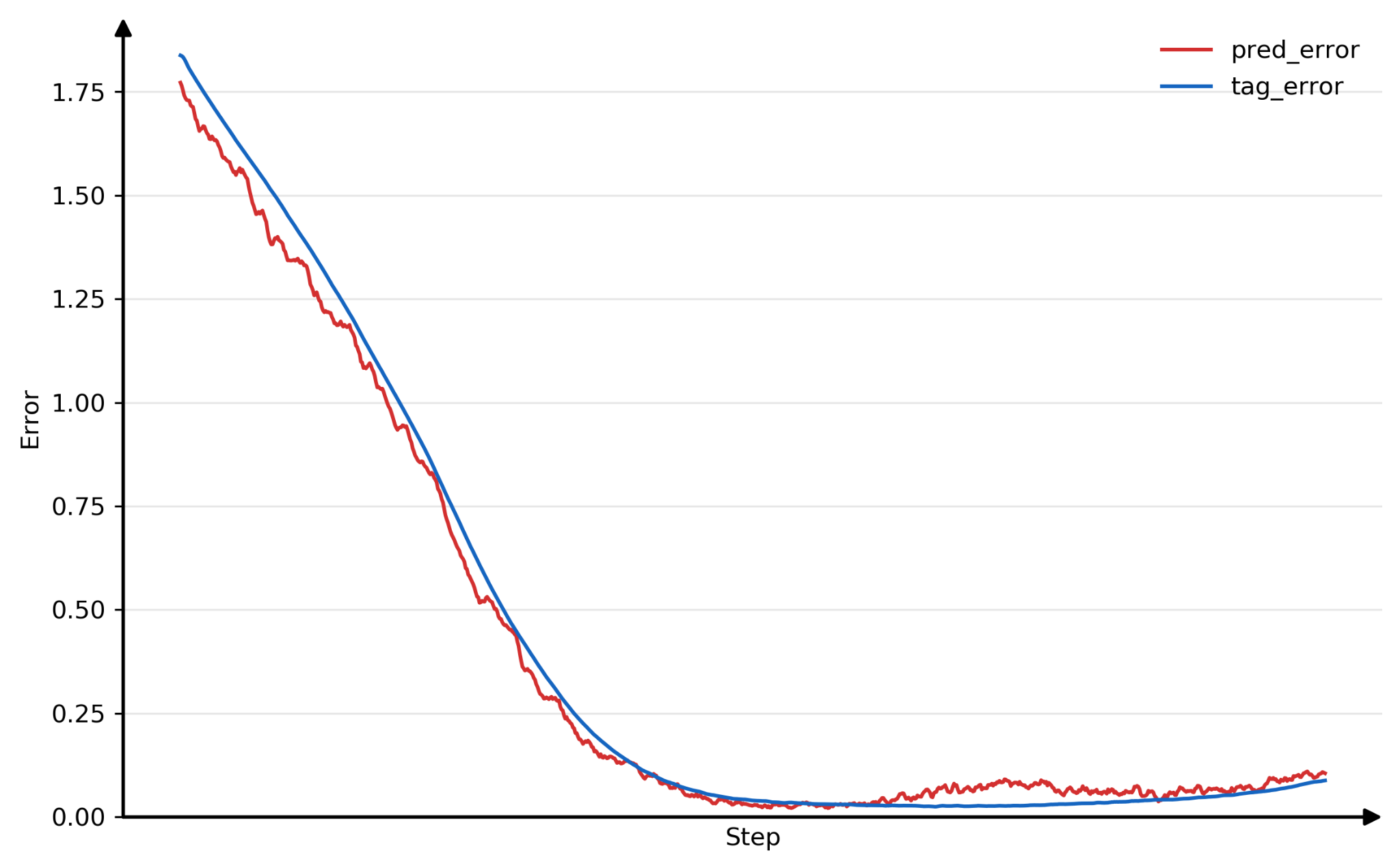}
\caption{\textbf{Consistency between learned progress and an external normalized AprilTag corner error.}
One example WM-VS trajectory is shown.
The red curve is the model-predicted visual error and the blue curve is $E_{\mathrm{tag}}$, the external normalized AprilTag corner error in (\ref{eq:tag_norm}).
The reported correlations are mean per-trajectory statistics over 30 real-robot trials.}
\label{fig:error_consistency}
\vspace{-0.25cm}
\end{figure}

Together, the correlation, ablation, and retention results link the Stage~1 representation to the closed-loop behavior of the final policy.

\subsection{Background Interference Evaluation}

\begin{figure}[t]
\centering
\setlength{\tabcolsep}{2pt}
\begin{tabular}{cc}
\includegraphics[width=0.48\linewidth]{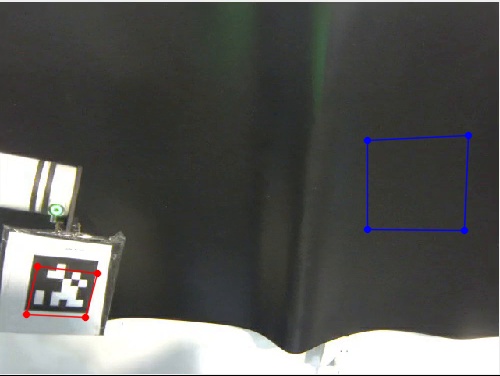} &
\includegraphics[width=0.48\linewidth]{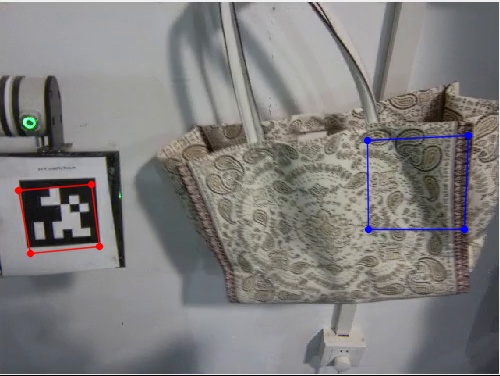} \\
{\small (a) Static black background} &
{\small (b) Dynamic background}
\end{tabular}
\caption{\textbf{Background interference evaluation.}
Representative best observed frames are shown for (a) a static black background and (b) a moving bag. The blue and red quadrilaterals mark the AprilTag corner detections at the initial and desired configurations.}
\label{fig:background_interference}
\vspace{-0.5cm}
\end{figure}

We also test background variation while keeping the servo target unchanged. A black wall gives the static condition and a patterned bag moving through the background gives the dynamic condition. Across 10 trial groups, convergence and final precision remain close to the nominal setting without a consistent degradation trend. Fig.~\ref{fig:background_interference} shows representative best observed frames under both conditions.

\subsection{Zero Shot Transfer to New 3D Targets}
\label{sec:zero_shot}

\begin{figure}[t]
\centering
\setlength{\tabcolsep}{2pt}
\begin{tabular}{cc}
\includegraphics[width=0.48\linewidth]{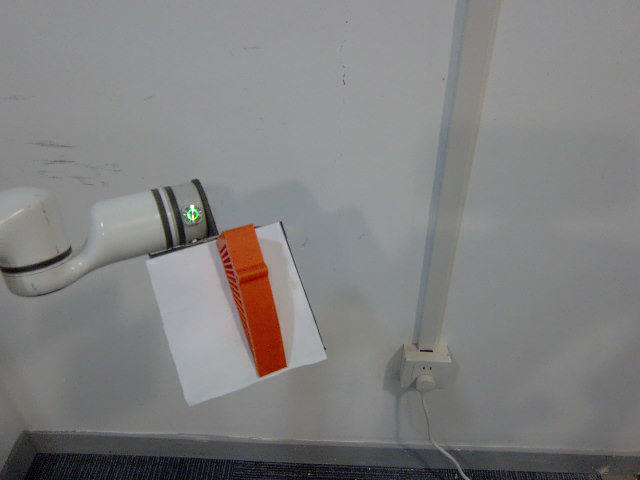} &
\includegraphics[width=0.48\linewidth]{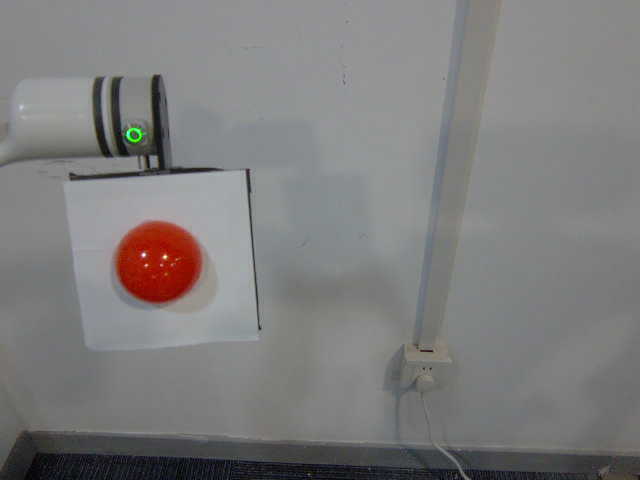} \\
{\small (a) Silicone gripper finger} &
{\small (b) Plastic hemisphere}
\end{tabular}
\caption{\textbf{Zero shot transfer to new 3D targets.} The two targets differ in shape and surface material.}
\label{fig:zero_shot}
\vspace{-0.5cm}
\end{figure}

We further test zero shot transfer on two 3D targets with different shapes and materials. The first is a silicone gripper finger and the second is a plastic hemisphere, shown in Fig.~\ref{fig:zero_shot}. The trained WM-VS controller is applied without additional task specific training.

At the best observed frame, the TCP translation and rotation errors are
\begin{equation}
\begin{aligned}
E_{\mathrm{trans}}
&=
\left\|\mathbf{p}_{\mathrm{best}}-\mathbf{p}_{g}\right\|_2,\\
E_{\mathrm{rot}}
&=
\cos^{-1}\!\left(
\frac{\mathrm{tr}\!\left(\mathbf{R}_{g}^{\top}\mathbf{R}_{\mathrm{best}}\right)-1}{2}
\right)\frac{180}{\pi}.
\end{aligned}
\label{eq:zero_shot_pose}
\end{equation}
The translation and rotation reduction columns in Table~\ref{tab:zero_shot} report the percentage decrease from the initial TCP error to the best observed TCP error.

\begin{table}[t]
\caption{Zero shot transfer results.}
\vspace{-0.25cm}
\label{tab:zero_shot}
\centering
\scriptsize
\setlength{\tabcolsep}{1.2pt}
\begin{tabular}{@{}lcccc@{}}
\toprule
Target & $E_{\mathrm{trans}}$ & $E_{\mathrm{rot}}$ & Trans. red. & Rot. red.\\
 & (mm) & (deg) & (\%) & (\%)\\
\midrule
Silicone finger & $45.66\pm19.46$ & $10.34\pm10.27$ & $86.48\pm5.85$ & $70.01\pm23.83$\\
Plastic hemisphere & $31.88\pm13.72$ & $9.76\pm5.89$ & $90.27\pm4.34$ & $65.70\pm20.03$\\
\bottomrule
\end{tabular}
\vspace{-0.5cm}
\end{table}

Both targets show clear closed-loop error reduction. Translation is reduced by $86.48\pm5.85\%$ and $90.27\pm4.34\%$, while rotation is reduced by $70.01\pm23.83\%$ and $65.70\pm20.03\%$ for the silicone finger and plastic hemisphere, respectively. Interestingly, RANSAC retains an average of $11.5$ correspondences on these unseen targets, compared with $8.0$ on the AprilTag target. The reduced zero-shot precision is therefore not explained simply by correspondence scarcity; the spatial conditioning and rotational informativeness of those correspondences can remain target dependent, particularly for smooth or nearly symmetric geometry.

\section{DISCUSSION}

WM-VS makes predicted action consequences measurable in target-relative progress rather than relying on action imitation alone. The ablations show that removing future-error alignment or Stage~2 consequence objectives preserves much of the reaching ability but sharply reduces final retention, linking progress alignment to post-convergence behavior. The external AprilTag correlation and background tests provide complementary checks. Unseen targets yield more surviving RANSAC correspondences than the AprilTag target, suggesting that transfer is limited by correspondence geometry and observability rather than count alone; broader target diversity is therefore a motivated direction for improving transfer.

\section{CONCLUSION}

We presented WM-VS for closed-loop visual servoing. It aligns predicted action consequences with a directional servo coordinate and trains a reactive policy for repeated feedback. WM-VS reaches the $10\%$ region in 30/30 trials and retains it in 25/30, while removing future error alignment lowers retention to $26.67\%$. Zero shot tests also reduce TCP translation and rotation error on two new 3D targets.


\FloatBarrier
\bibliographystyle{IEEEtran}
\bibliography{references}

@article{hutchinson1996tutorial,
  author={Hutchinson, S. and Hager, G.D. and Corke, P.I.},
  journal={IEEE Transactions on Robotics and Automation}, 
  title={A tutorial on visual servo control}, 
  year={1996},
  volume={12},
  number={5},
  pages={651-670},
  doi={10.1109/70.538972}}

@article{chaumette2006visual,
  author={Chaumette, Francois and Hutchinson, Seth},
  journal={IEEE Robotics \& Automation Magazine}, 
  title={Visual servo control. I. Basic approaches}, 
  year={2006},
  volume={13},
  number={4},
  pages={82-90},
  doi={10.1109/MRA.2006.250573}}

@article{chaumette2007visual,
  author={Chaumette, Francois and Hutchinson, Seth},
  journal={IEEE Robotics \& Automation Magazine}, 
  title={Visual servo control. II. Advanced approaches [Tutorial]}, 
  year={2007},
  volume={14},
  number={1},
  pages={109-118},
  doi={10.1109/MRA.2007.339609}}

@article{chaumette2004moments,
  author={Chaumette, F.},
  journal={IEEE Transactions on Robotics}, 
  title={Image moments: a general and useful set of features for visual servoing}, 
  year={2004},
  volume={20},
  number={4},
  pages={713-723},
  doi={10.1109/TRO.2004.829463}}

@article{marchand2005visp,
  author={Marchand, E. and Spindler, F. and Chaumette, F.},
  journal={IEEE Robotics \& Automation Magazine}, 
  title={{ViSP} for visual servoing: a generic software platform with a wide class of robot control skills}, 
  year={2005},
  volume={12},
  number={4},
  pages={40-52},
  doi={10.1109/MRA.2005.1577023}}

@inproceedings{collewet2008setfree,
  author = {Christophe Collewet and Eric Marchand and Fran{\c{c}}ois Chaumette},
  booktitle={2008 IEEE International Conference on Robotics and Automation}, 
  title={Visual servoing set free from image processing}, 
  year={2008},
  volume={},
  number={},
  pages={81-86},
  doi={10.1109/ROBOT.2008.4543190}}

@article{collewet2011photometric,
  author={Collewet, Christophe and Marchand, Eric},
  journal={IEEE Transactions on Robotics}, 
  title={Photometric Visual Servoing}, 
  year={2011},
  volume={27},
  number={4},
  pages={828-834},
  doi={10.1109/TRO.2011.2112593}}

@inproceedings{lazar2009mpc,
  author={Lazar, C. and Burlacu, A.},
  booktitle={2009 7th IEEE International Conference on Industrial Informatics}, 
  title={Visual Servoing of Robot Manipulators Using Model-based Predictive Control}, 
  year={2009},
  volume={},
  number={},
  pages={690-695},
  doi={10.1109/INDIN.2009.5195887}}

@article{allibert2010predictive,
  author={Allibert, Guillaume and Courtial, Estelle and Chaumette, François},
  journal={IEEE Transactions on Robotics}, 
  title={Predictive Control for Constrained Image-Based Visual Servoing}, 
  year={2010},
  volume={26},
  number={5},
  pages={933-939},
  doi={10.1109/TRO.2010.2056590}}

@article{wu2022survey,
    title = {A survey Of learning-Based control of robotic visual servoing systems},
    journal = {Journal of the Franklin Institute},
    volume = {359},
    number = {1},
    pages = {556-577},
    year = {2022},
    issn = {0016-0032},
    doi = {https://doi.org/10.1016/j.jfranklin.2021.11.009},
    url = {https://www.sciencedirect.com/science/article/pii/S0016003221006621},
    author = {Jinhui Wu and Zhehao Jin and Andong Liu and Li Yu and Fuwen Yang}
    }

@inproceedings{saxena2017exploring,
  author={Saxena, Aseem and Pandya, Harit and Kumar, Gourav and Gaud, Ayush and Krishna, K. Madhava},
  booktitle={2017 IEEE International Conference on Robotics and Automation (ICRA)}, 
  title={Exploring convolutional networks for end-to-end visual servoing}, 
  year={2017},
  volume={},
  number={},
  pages={3817-3823},
  doi={10.1109/ICRA.2017.7989442}}

@inproceedings{bateux2018training,
  author={Bateux, Quentin and Marchand, Eric and Leitner, J{\"u}rgen and Chaumette, Fran{\c{c}}ois and Corke, Peter},
  booktitle={2018 IEEE International Conference on Robotics and Automation (ICRA)}, 
  title={Training Deep Neural Networks for Visual Servoing}, 
  year={2018},
  volume={},
  number={},
  pages={3307-3314},
  doi={10.1109/ICRA.2018.8461068}}

@inproceedings{yu2019siamese,
  author={Yu, Cunjun and Cai, Zhongang and Pham, Hung and Pham, Quang-Cuong},
  booktitle={2019 IEEE/RSJ International Conference on Intelligent Robots and Systems (IROS)}, 
  title={Siamese Convolutional Neural Network for Sub-millimeter-accurate Camera Pose Estimation and Visual Servoing}, 
  year={2019},
  volume={},
  number={},
  pages={935-941},
  doi={10.1109/IROS40897.2019.8967925}}

@inproceedings{lee2017learning,
    title={Learning Visual Servoing with Deep Features and Fitted Q-Iteration},
    author={Alex X. Lee and Sergey Levine and Pieter Abbeel},
    booktitle={International Conference on Learning Representations},
    year={2017}
    }

@inproceedings{sampedro2018deeprl,
  author={Sampedro, Carlos and Rodriguez-Ramos, Alejandro and Gil, Ignacio and Mejias, Luis and Campoy, Pascual},
  booktitle={2018 IEEE/RSJ International Conference on Intelligent Robots and Systems (IROS)}, 
  title={Image-Based Visual Servoing Controller for Multirotor Aerial Robots Using Deep Reinforcement Learning}, 
  year={2018},
  volume={},
  number={},
  pages={979-986},
  doi={10.1109/IROS.2018.8594249}}

@inproceedings{harish2020dfvs,
  author={Harish, Y V S and Pandya, Harit and Gaud, Ayush and Terupally, Shreya and Shankar, Sai and Krishna, K. Madhava},
  booktitle={2020 IEEE International Conference on Robotics and Automation (ICRA)}, 
  title={{DFVS}: Deep Flow Guided Scene Agnostic Image Based Visual Servoing}, 
  year={2020},
  volume={},
  number={},
  pages={9000-9006},
  doi={10.1109/ICRA40945.2020.9196753}}

@inproceedings{katara2021deepmpcvs,
  title = 	 {{DeepMPCVS}: Deep Model Predictive Control for Visual Servoing},
  author =       {Katara, Pushkal and YVS, Harish and Pandya, Harit and Gupta, Abhinav and Sanchawala, AadilMehdi and Kumar, Gourav and Bhowmick, Brojeshwar and Krishna, Madhava},
  booktitle = 	 {Proceedings of the 2020 Conference on Robot Learning},
  pages = 	 {2006--2015},
  year = 	 {2021},
  editor = 	 {Kober, Jens and Ramos, Fabio and Tomlin, Claire},
  volume = 	 {155},
  series = 	 {Proceedings of Machine Learning Research},
  month = 	 {16--18 Nov},
  publisher =    {PMLR},
  url = 	 {https://proceedings.mlr.press/v155/katara21a.html}
}

@inproceedings{felton2021siamese3,
  author={Allibert, Guillaume and Courtial, Estelle and Chaumette, Fran{\c{c}}ois},
  booktitle={2021 IEEE International Conference on Robotics and Automation (ICRA)}, 
  title={Siame-se(3): regression in se(3) for end-to-end visual servoing}, 
  year={2021},
  volume={},
  number={},
  pages={14454-14460},
  doi={10.1109/ICRA48506.2021.9561488}}

@inproceedings{puang2020kovis,
  author={Puang, En Yen and Peng Tee, Keng and Jing, Wei},
  booktitle={2020 IEEE/RSJ International Conference on Intelligent Robots and Systems (IROS)}, 
  title={{KOVIS}: Keypoint-based Visual Servoing with Zero-Shot Sim-to-Real Transfer for Robotics Manipulation}, 
  year={2020},
  volume={},
  number={},
  pages={7527-7533},
  doi={10.1109/IROS45743.2020.9341370}}

@inproceedings{adrian2022dfbvs,
  author={Adrian, Nicholas and Do, Van-Thach and Pham, Quang-Cuong},
  booktitle={2022 IEEE 18th International Conference on Automation Science and Engineering (CASE)}, 
  title={{DFBVS}: Deep Feature-Based Visual Servo}, 
  year={2022},
  volume={},
  number={},
  pages={1783-1789},
  doi={10.1109/CASE49997.2022.9926560}}

@article{felton2022autoencoder,
  author={Felton, Samuel and Brault, Pascal and Fromont, Elisa and Marchand, Eric},
  journal={IEEE Robotics and Automation Letters}, 
  title={Visual Servoing in Autoencoder Latent Space}, 
  year={2022},
  volume={7},
  number={2},
  pages={3234-3241},
  doi={10.1109/LRA.2022.3144490}}

@inproceedings{haugaard2022self,
  author={Haugaard, Rasmus Laurvig and Glent Buch, Anders and Iversen, Thorbj{\o}rn Mosekj{\ae}r},
  booktitle={2022 IEEE 18th International Conference on Automation Science and Engineering (CASE)}, 
  title={Self-supervised deep visual servoing for high precision peg-in-hole insertion}, 
  year={2022},
  volume={},
  number={},
  pages={405-410},
  doi={10.1109/CASE49997.2022.9926468}}

@inproceedings{felton2023dmlvs,
  TITLE = {{Deep metric learning for visual servoing: when pose and image meet in latent space}},
  AUTHOR = {Felton, Samuel and Fromont, {\'E}lisa and Marchand, Eric},
  URL = {https://inria.hal.science/hal-04003126},
  BOOKTITLE = {{ICRA 2023 - IEEE International Conference on Robotics and Automation}},
  ADDRESS = {London, United Kingdom},
  PUBLISHER = {{IEEE}},
  PAGES = {741-747},
  YEAR = {2023},
  MONTH = May,
  DOI = {10.1109/ICRA48891.2023.10160963},
  HAL_ID = {hal-04003126},
  HAL_VERSION = {v1},
}

@inproceedings{scherl2025vitvs,
  author={Scherl, Alessandro and Thalhammer, Stefan and Neuberger, Bernhard and W{\"o}ber, Wilfried and Garc{\'i}a-Rodr{\'i}guez, Jos{\'e}},
  booktitle={2025 IEEE/RSJ International Conference on Intelligent Robots and Systems (IROS)}, 
  title={{ViT-VS}: On the Applicability of Pretrained Vision Transformer Features for Generalizable Visual Servoing}, 
  year={2025},
  volume={},
  number={},
  pages={17769-17776},
  doi={10.1109/IROS60139.2025.11246608}}

@inproceedings{dosovitskiy2021vit,
      title={An Image is Worth 16x16 Words: Transformers for Image Recognition at Scale}, 
      author={Alexey Dosovitskiy and Lucas Beyer and Alexander Kolesnikov and Dirk Weissenborn and Xiaohua Zhai and Thomas Unterthiner and Mostafa Dehghani and Matthias Minderer and Georg Heigold and Sylvain Gelly and Jakob Uszkoreit and Neil Houlsby},
      booktitle={Proceedings of the International Conference on Learning Representations (ICLR)},
      year={2021},
      eprint={2010.11929},
      archivePrefix={arXiv},
      primaryClass={cs.CV},
      url={https://arxiv.org/abs/2010.11929}, 
}

@inproceedings{caron2021dino,
  author={Caron, Mathilde and Touvron, Hugo and Misra, Ishan and Jegou, Herv{\'e} and Mairal, Julien and Bojanowski, Piotr and Joulin, Armand},
  booktitle={2021 IEEE/CVF International Conference on Computer Vision (ICCV)}, 
  title={Emerging Properties in Self-Supervised Vision Transformers}, 
  year={2021},
  volume={},
  number={},
  pages={9630-9640},
  doi={10.1109/ICCV48922.2021.00951}}

@article{oquab2024dinov2,
    title={{DINO}v2: Learning Robust Visual Features without Supervision},
    author={Maxime Oquab and Timoth{\'e}e Darcet and Th{\'e}o Moutakanni and Huy V. Vo and Marc Szafraniec and Vasil Khalidov and Pierre Fernandez and Daniel HAZIZA and Francisco Massa and Alaaeldin El-Nouby and Mido Assran and Nicolas Ballas and Wojciech Galuba and Russell Howes and Po-Yao Huang and Shang-Wen Li and Ishan Misra and Michael Rabbat and Vasu Sharma and Gabriel Synnaeve and Hu Xu and Herve Jegou and Julien Mairal and Patrick Labatut and Armand Joulin and Piotr Bojanowski},
    journal={Transactions on Machine Learning Research},
    issn={2835-8856},
    year={2024},
    note={Featured Certification}
}

@inproceedings{amir2022deepvit,
      title={Deep {ViT} Features as Dense Visual Descriptors}, 
      author={Shir Amir and Yossi Gandelsman and Shai Bagon and Tali Dekel},
      year={2022},
      eprint={2112.05814},
      archivePrefix={arXiv},
      primaryClass={cs.CV},
      url={https://arxiv.org/abs/2112.05814},
      booktitle = {arXiv preprint},
}

@inproceedings{dekel2015bestbuddies,
    author = {Dekel, Tali and Oron, Shaul and Rubinstein, Michael and Avidan, Shai and Freeman, William T.},
    title = {Best-Buddies Similarity for Robust Template Matching},
    booktitle = {Proceedings of the IEEE Conference on Computer Vision and Pattern Recognition (CVPR)},
    month = {June},
    year = {2015}
}

@inproceedings{kirillov2023sam,
  author={Kirillov, Alexander and Mintun, Eric and Ravi, Nikhila and Mao, Hanzi and Rolland, Chloe and Gustafson, Laura and Xiao, Tete and Whitehead, Spencer and Berg, Alexander C. and Lo, Wan-Yen and Doll{\'a}r, Piotr and Girshick, Ross},
  booktitle={2023 IEEE/CVF International Conference on Computer Vision (ICCV)}, 
  title={Segment Anything}, 
  year={2023},
  volume={},
  number={},
  pages={3992-4003},
  doi={10.1109/ICCV51070.2023.00371}}

@inproceedings{cheng2022mask2former,
  author={Cheng, Bowen and Misra, Ishan and Schwing, Alexander G. and Kirillov, Alexander and Girdhar, Rohit},
  booktitle={2022 IEEE/CVF Conference on Computer Vision and Pattern Recognition (CVPR)}, 
  title={Masked-attention Mask Transformer for Universal Image Segmentation}, 
  year={2022},
  volume={},
  number={},
  pages={1280-1289},
  doi={10.1109/CVPR52688.2022.00135}}

@inproceedings{finn2017foresight,
  author={Finn, Chelsea and Levine, Sergey},
  booktitle={2017 IEEE International Conference on Robotics and Automation (ICRA)}, 
  title={Deep visual foresight for planning robot motion}, 
  year={2017},
  volume={},
  number={},
  pages={2786-2793},
  doi={10.1109/ICRA.2017.7989324}}

@article{hirose2019deepvisualmpc,
  author={Hirose, Noriaki and Xia, Fei and Mart{\'i}n-Mart{\'i}n, Roberto and Sadeghian, Amir and Savarese, Silvio},
  journal={IEEE Robotics and Automation Letters}, 
  title={Deep Visual MPC-Policy Learning for Navigation}, 
  year={2019},
  volume={4},
  number={4},
  pages={3184-3191},
  doi={10.1109/LRA.2019.2925731}}

@misc{hafner2020dreamer,
    title={Dream to Control: Learning Behaviors by Latent Imagination},
    author={Danijar Hafner and Timothy Lillicrap and Jimmy Ba and Mohammad Norouzi},
    booktitle={International Conference on Learning Representations},
    year={2020}
}

@inproceedings{olson2011apriltag,
  author={Olson, Edwin},
  booktitle={2011 IEEE International Conference on Robotics and Automation}, 
  title={{AprilTag}: A robust and flexible visual fiducial system}, 
  year={2011},
  volume={},
  number={},
  pages={3400-3407},
  doi={10.1109/ICRA.2011.5979561}}

@article{ma2026uni,
  author={Ma, Junyi and Bao, Wentao and Xu, Jingyi and Sun, Guanzhong and Zheng, Yu and Zhang, Erhang and Chen, Xieyuanli and Wang, Hesheng},
  journal={IEEE Transactions on Pattern Analysis and Machine Intelligence}, 
  title={Uni-Hand: Universal Hand Motion Forecasting in Egocentric Views}, 
  year={2026},
  volume={48},
  number={10},
  pages={11758-11775},
  doi={10.1109/TPAMI.2026.3692504}}

\end{document}